\documentclass[10pt,twocolumn,letterpaper]{article}

\usepackage{wacv}              

\usepackage{wrapfig}
\usepackage{overpic}
\usepackage{enumitem} 
\usepackage{color}
\usepackage{comment}
\usepackage{booktabs} 
\usepackage[accsupp]{axessibility} 

\usepackage[ruled, vlined]{algorithm2e}
\SetAlFnt{\small}     
\SetAlCapFnt{\small}   

\usepackage{amsfonts}
\usepackage{threeparttable}
\usepackage{tabularx}
\usepackage{mathtools}

\usepackage{caption}
\usepackage{subcaption}
\usepackage{multirow}
\usepackage{soul}
\usepackage[export]{adjustbox}
\usepackage{multirow}
\usepackage{arydshln} 
\usepackage{stmaryrd}
\usepackage{makecell}
\usepackage{tabu}
\usepackage{diagbox} 
\usepackage{siunitx}   
\newcolumntype{L}{>{\raggedright\arraybackslash}X} 
\newcolumntype{C}{>{\centering\arraybackslash}X}  
\usepackage[hang,flushmargin]{footmisc} 
\usepackage{array}

\usepackage{pgfplots}
\pgfplotsset{width=10cm,compat=1.9}

\usepgfplotslibrary{external}
\usepackage{amsmath,amsfonts,bm,mathtools}

\def\eqref#1{equation~\ref{#1}}

\def\1{\bm{1}}

\DeclareMathAlphabet{\mathsfit}{\encodingdefault}{\sfdefault}{m}{sl}
\SetMathAlphabet{\mathsfit}{bold}{\encodingdefault}{\sfdefault}{bx}{n}

\def\gM{{\mathcal{M}}}

\def\sR{{\mathbb{R}}}

\newcommand{\N}{\mathbb{N}}

\usepackage{microtype} 
\usepackage{placeins} 

\definecolor{turquoise}{cmyk}{0.65,0,0.1,0.3}
\definecolor{purple}{rgb}{0.65,0,0.65}
\definecolor{dark_green}{rgb}{0, 0.5, 0}
\definecolor{orange}{rgb}{0.8, 0.6, 0.2}
\definecolor{red}{rgb}{0.9, 0.1, 0.1}
\definecolor{darkred}{rgb}{0.6, 0.1, 0.05}
\definecolor{blueish}{rgb}{0.0, 0.3, .6}
\definecolor{light_gray}{rgb}{0.7, 0.7, .7}
\definecolor{pink}{rgb}{1, 0, 1}
\definecolor{greyblue}{rgb}{0.25, 0.25, 1}
\definecolor{commentgray}{gray}{0.6}

\newcommand{\methodname}{MSD\xspace}

\def\N{{\mathcal{N}}}

\newcommand\blfootnote[1]{%
  \begingroup
  \renewcommand\thefootnote{}\footnote{#1}%
  \addtocounter{footnote}{-1}%
  \endgroup
}

\usepackage{blindtext}

\ifpdf

\else
\fi

\ifpdf

\else
\fi

\ifpdf

\else
\fi

\usepackage{color}
\definecolor{cyan}{cmyk}{1,0,0,0}
\definecolor{darkgreen}{rgb}{0,0.5,0}
\definecolor{darkyellow}{rgb}{0.96,0.61,0.25}
\definecolor{darkred}{rgb}{0.9,0.1,0.1}
\definecolor{orange}{rgb}{1,0.5,0}
\definecolor{magenta}{cmyk}{0,1,0,0}
\definecolor{gray}{rgb}{0.8,0.8,0.8}
\definecolor{good}{rgb}{0.75,0.9,0.75}
\definecolor{decent}{rgb}{0.9,0.93,0.75}
\definecolor{bad}{rgb}{0.9,0.75,0.75}
\definecolor{na}{rgb}{0.8,0.8,0.8}
\definecolor{textprompts}{rgb}{0.0,0.705,0.313}

\definecolor{fontgreen}{RGB}{0,190,0}
\definecolor{glyphblue}{RGB}{0,0,220}

\SetKwComment{Comment}{\color{green!50!black}// }{}

\SetKwProg{Function}{function}{}{}

\definecolor{gold}{RGB}{255,215,0}
\definecolor{silver}{RGB}{192,192,192}

\newcommand{\hlgold}[1]{\sethlcolor{gold}\hl{#1}}
\newcommand{\hlsilver}[1]{\sethlcolor{silver}\hl{#1}}

\newcommand{\expkernel}{e^{- \Delta t / \lambda^2}}

\definecolor{wacvblue}{rgb}{0.21,0.49,0.74}
\usepackage[pagebackref,breaklinks,colorlinks,allcolors=wacvblue]{hyperref}

\def\wacvPaperID{1886} 
\def\confName{WACV}
\def\confYear{2026}

\title{Mean-Shift Distillation for Diffusion Mode Seeking}

\author{
Vikas Thamizharasan\\
{\normalsize UMass Amherst \& Adobe Research}\\
\and
Nikitas Chatzis\\
{\normalsize NTUA}\\
\and
Iliyan Georgiev\\
{\normalsize Adobe Research}\\
\and
Matthew Fisher\\
{\normalsize Adobe Research}\\
\and
Evangelos Kalogerakis\\
{\normalsize TU Crete \& UMass Amherst}\\
\and
Difan Liu\\
{\normalsize Adobe Research}\\
\and
Nanxuan Zhao\\
{\normalsize Adobe Research}\\
\and
Michal Lukáč\\
{\normalsize Adobe Research}\\
}

\begin{document}
\maketitle

\begin{abstract}
We present \emph{mean-shift distillation}, a novel diffusion distillation technique that provides a provably good proxy for the gradient of the diffusion output distribution. This is derived directly from mean-shift mode seeking on the distribution, and we show that its extrema are aligned with the modes. We further derive an efficient product distribution sampling procedure to evaluate the gradient. Our method is formulated as a drop-in replacement for score distillation sampling (SDS), requiring neither model retraining nor extensive modification of the sampling procedure. We show that it exhibits superior mode alignment as well as improved convergence in both synthetic and practical setups, yielding higher-fidelity results when applied to both text-to-image and text-to-3D applications with Stable Diffusion.
\end{abstract}

\blfootnote{
~Project website: \textcolor{red}{https://vikastmz.github.io/MSD/}
}

\vspace{-4mm}
\section{Introduction}
\label{sec:intro}

Soon after image diffusion~\cite{dhariwal2021diffusion} models exploded in popularity, DreamFusion \cite{poole2022dreamfusion} and SJC \cite{sjc} concurrently introduced the idea of using them for image optimization. Intuitively, this can be expressed as the notion that images more likely to be generated by a diffusion model are ``better'' in the sense of being more faithful to the data distribution the diffusion model was trained on.

Formally, diffusion models provide a mechanism to sample images $x \in \mathcal{I}$ from a learned distribution $p(x)$. We then have a parameter vector $\vartheta \in \mathcal{P}$, along with an image-generating model $g: \mathcal{P} \rightarrow \mathcal{I}$. Given an initialization $\vartheta^0$, we seek to optimize a $\vartheta^k$ such that $p(g(\vartheta^k)) > p(g(\vartheta^0))$. We expect this to yield an image $g(\vartheta^k)$ of higher quality, under the metric the diffusion model is trained for.


We could imagine optimizing $\vartheta$ by determining the gradient $\nabla p(g(\vartheta))$ and ascending along it. However, while we may use our diffusion model to sample from $p(x)$, we can neither easily evaluate $p(x)$ nor determine its gradient $\nabla_x p(x)$. Even though we can formally express $p(x)$ in terms of the score function $\epsilon(x,t)$ through the instantaneous change of variable formula~\cite{grathwohl2018scalable}, evaluating this formula requires calculating the divergence of the score function along the entire ODE path, making this of only theoretical interest. Evaluating the gradient of this quantity is even less practical.

Score distillation sampling (SDS)~\cite{poole2022dreamfusion,sjc} attempts to address this problem by offering proxies for the density gradient that are easier to estimate. However, their theoretical properties are not rigorously established, and SDS suffers from significant bias as well as variance, yielding inaccurate gradients. Examining the loss landscape of SDS in \cref{fig:fractal_2D_part1}, we indeed see that not only are the maxima of this function not collocated with the modes of $p(x)$, but even in the simplest cases the loss creates ``phantom modes'' that are well out of distribution. Our method offers both better alignment with the distribution and lower variance of the gradient estimate.

\vspace{-4mm}
\paragraph{Contributions.}

In this paper, we propose \emph{mean-shift distillation}, a distribution-gradient proxy based on a well-known mode-seeking technique. Furthermore, we show that:
\begin{itemize}
    \item This proxy can be implemented easily, with minimal changes to the diffusion sampling procedure;
    \item It evaluates with less variance than SDS with improved mode alignment;
    \item It has superior behavior, converging to modes of the trained distribution with a clear termination criterion.
\end{itemize}

\section{Related Work}
\label{sec:related}

\paragraph{Denoising diffusion.}

In our work we rely most directly on the mean-shift method of mode seeking~\cite{Cheng98,ComaniciuM02Meanshift}, but our ability to apply it to diffusion rests on a body of theoretical analysis of this process.

Mathematically, denoising diffusion consists of solving an initial value problem (IVP) on a random variable from a simple, typically standard normal distribution, where the time-dependent gradient is learned by reversing the process of adding noise to the distribution being modeled~\cite{song2021scorebased,song2021denoising}. Already in these works authors suggest ways in which the output distribution may be manipulated by adding terms to the differential equation underlying the initial value problem, a property we will rely on to manipulate the output to our method's advantage.

A surprising connection between mean shift and diffusion emerged from the analysis of the optimal denoising model~\cite{Karras2022edm,chen2024a, symmetryJihyeon2024}. Since the forward (noising) process can be expressed as successive convolutions with a Gaussian kernel, the intermediate distributions are in fact Gaussian-kernel density estimates of the data distributions, with kernel bandwidth proportional to the time parameter. Therefore in the ODE of the reverse (inference) process, the gradient of the denoiser is theoretically equal to the mean-shift vector with appropriate kernel and bandwidth. Mean-shifting on the IVP time domain does not in fact seek modes of the output distribution, but we take advantage of this knowledge to implement mean shift on that domain.

Further related to the analysis of modes in particular, \cite{karras2024guiding, bradley2024classifierfreeguidancepredictorcorrector} suggest that applying classifier-free guidance (CFG)~\cite{ho2021classifierfree} to diffusion has the effect of sharpening the modes of the output distribution. This guidance does not explicitly \emph{seek} modes, but we have found that using CFG synergizes well with both SDS and our method.

\vspace{-4mm}
\paragraph{Distilling diffusion priors.}

Score distillation sampling (SDS) \cite{poole2022dreamfusion,sjc} has emerged as a useful technique for leveraging the priors learned by large-scale image models beyond 2D raster images. SDS provides an optimization procedure to estimate the parameters of a differentiable image generator, such that the rendered image is pushed towards a higher-probability region of a pre-trained prompt-conditioned image diffusion model. Originally proposed to optimize volumetric representations like NeRFs, it has been extended to other non-pixel-based representations \cite{jain2023vectorfusion,yi2023gaussiandreamer,bah20244dfy,Thamizharasan_2024_CVPR}.
    
The tendency of SDS to produce over-smoothened results due to high variance is well documented. A plethora of works have been proposed to mitigate this behavior, e.g.\ to factorize the gradient to reduce the bias \cite{hertz2023delta,yu2024texttod,katzir2024noisefree,alldieck2024scoredistillationsamplinglearned}, or
to replace the uniform noise sampling in SDS with noise obtained by running DDIM inversion \cite{EnVision2023luciddreamer,lukoianov2024score}. SteinDreamer \cite{wang2023steindreamer} propose a control variate for SDS, \cite{xu2024diversescoredistillation,yan2025consistentflowdistillation} improve diversity of generations, and \cite{wang2024esd} alleviate the multi-view inconsistency problem.

VSD \cite{wang2023prolificdreamer} tackle the low-fidelity problem by treating the target parameters as a random variable and estimate the variational distribution to produce diverse, high-fidelity results. SDI \cite{lukoianov2024score}, on the other hand and unlike previous variance-reduction methods, find a better approximation of the added noise term, eliminating one of the root causes of the excessive variance. These methods attribute SDS/SJC to be mode-seeking; we show it is not and introduce mode-seeking behavior to address the excessive variance and low-fidelity results.

We draw the distinction between the above methods to knowledge distillation works designed for one-step inference \cite{yin2024onestep, luo2023diffinstruct, xie2024em}. Diff-Instruct \cite{luo2023diffinstruct} show that SDS is a special case of their distillation formulation when the generator’s output is a Dirac’s Delta distribution and the marginal of the variational score is Gaussian. While the derived gradients resemble SDS and VSD, these methods require training an auxiliary score network to estimate the variational score--- challenging to generalize beyond text-to-image generation--- and have been targeted for different use cases, namely faster inference, model compression, and dataset privacy-preserving.

\newcommand{\datasample}{\textcolor{black}{$\bullet$}}
\newcommand{\withguidance}{\textcolor{darkgreen}{$\bigstar$}}
\newcommand{\noguidance}{\textcolor{darkred}{$\otimes$}}
\newcommand{\guidancelimitedinterval}{\textcolor{cyan}{$\blacklozenge$}}

\newcommand{\classone}{\textcolor{orange}{orange}}
\newcommand{\classtwo}{\textcolor{gray}{gray}}

\begin{figure*}[t]
    \centering
    \setlength{\tabcolsep}{1.4mm}
    \begin{tabular}{c c c c}
        \includegraphics[width=0.2\textwidth]{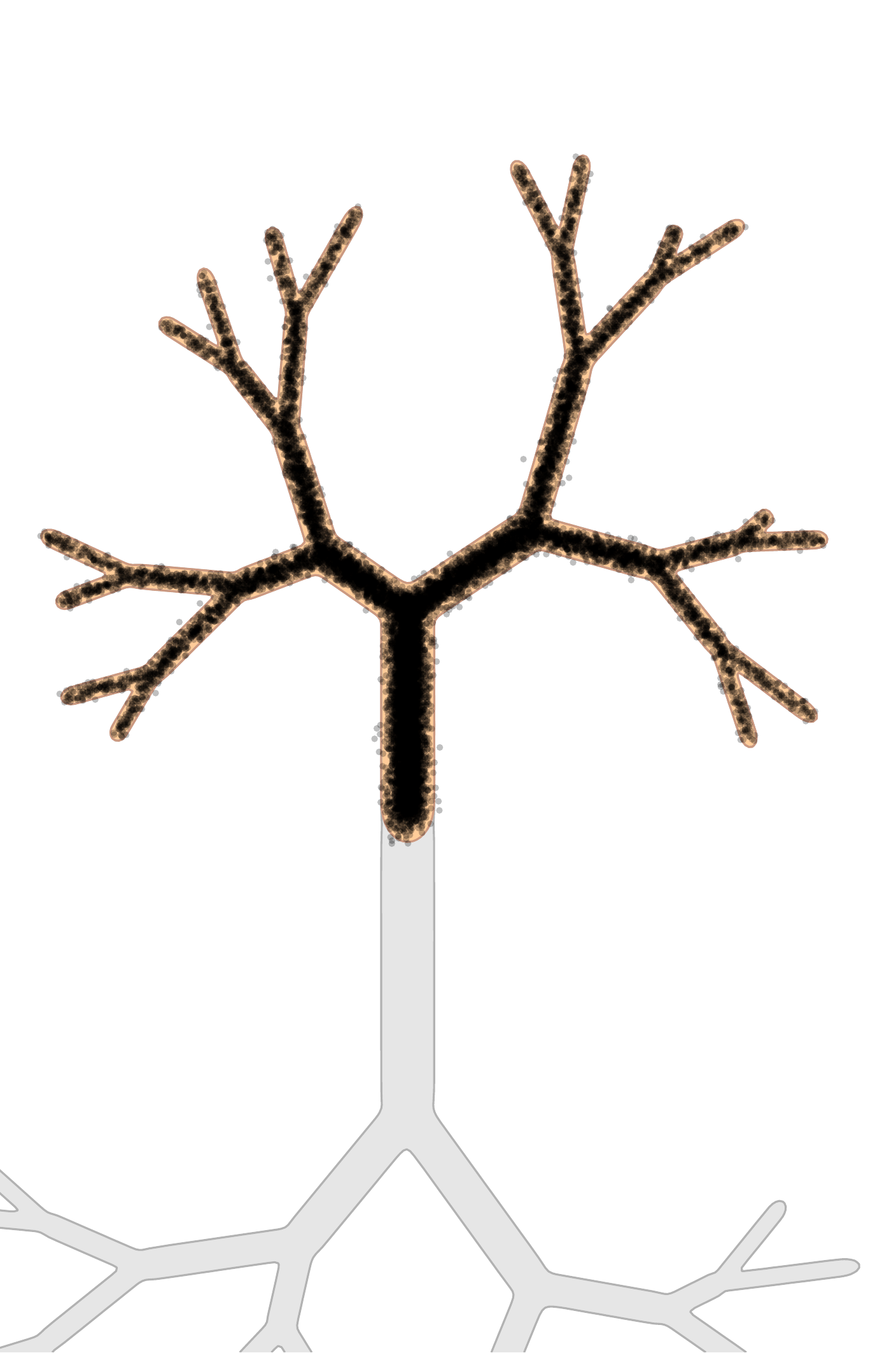}
        & 
        \includegraphics[width=0.2\textwidth]{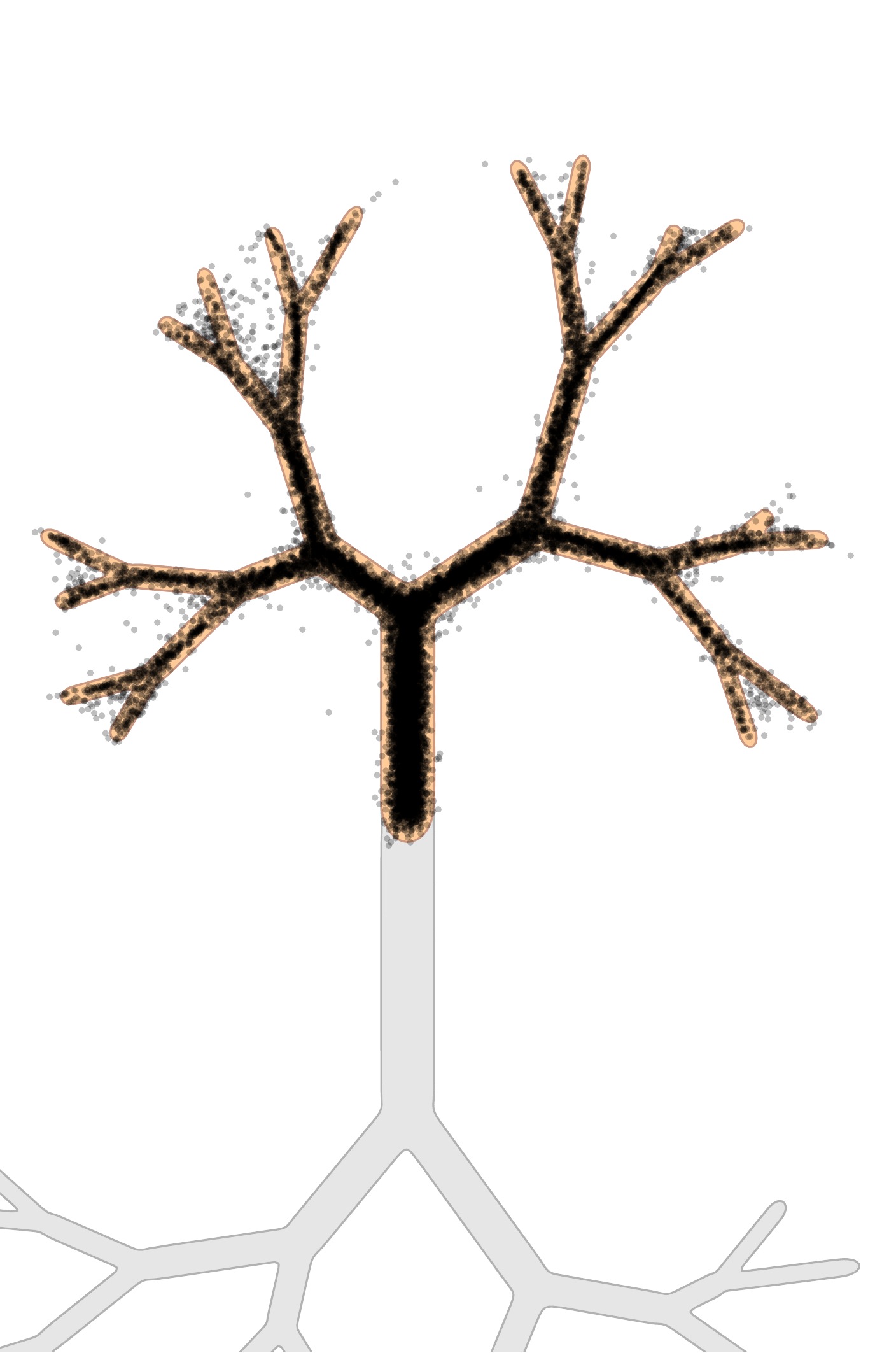}
        &
        \includegraphics[width=0.2\textwidth]{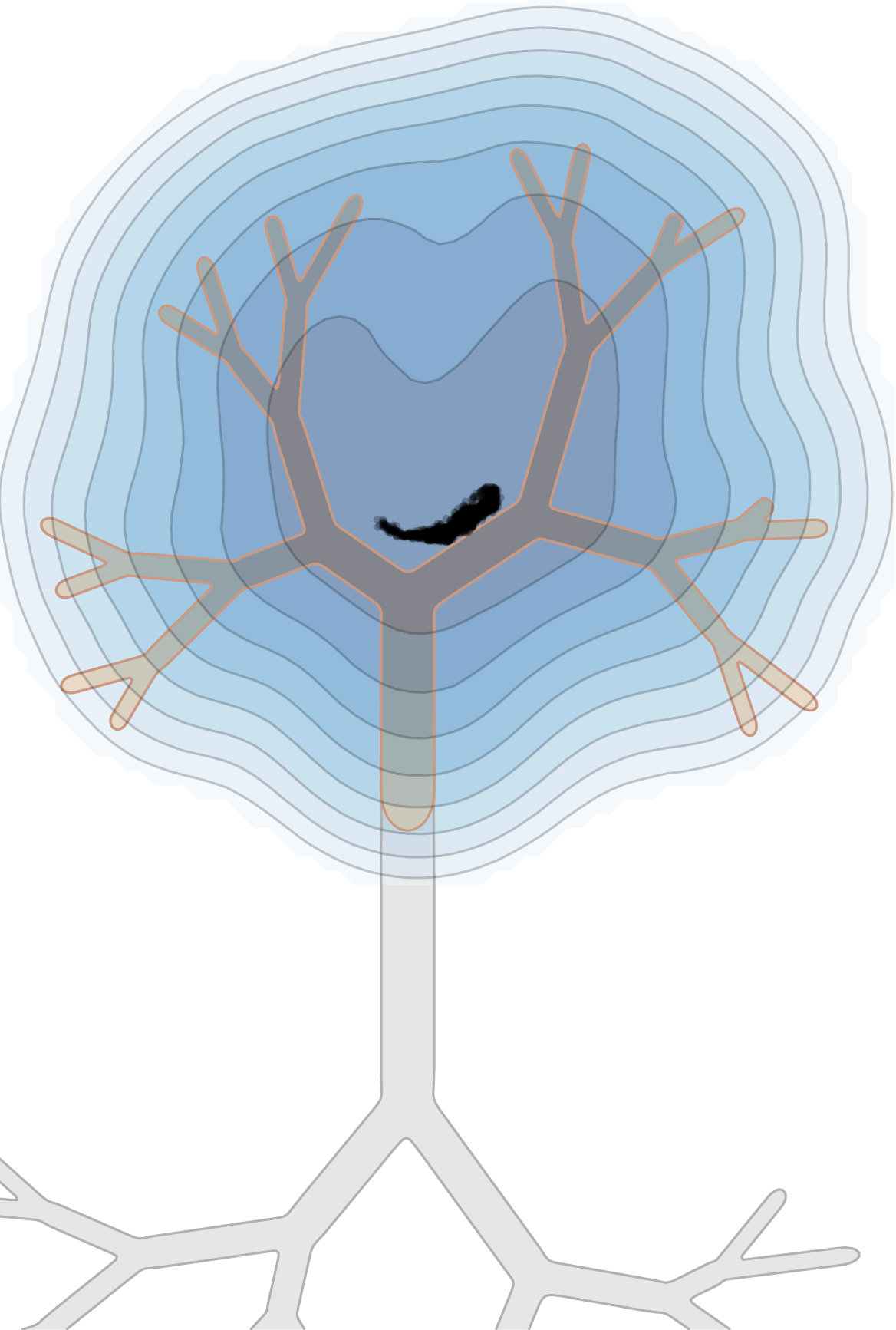}
        &
        \includegraphics[width=0.2\textwidth]{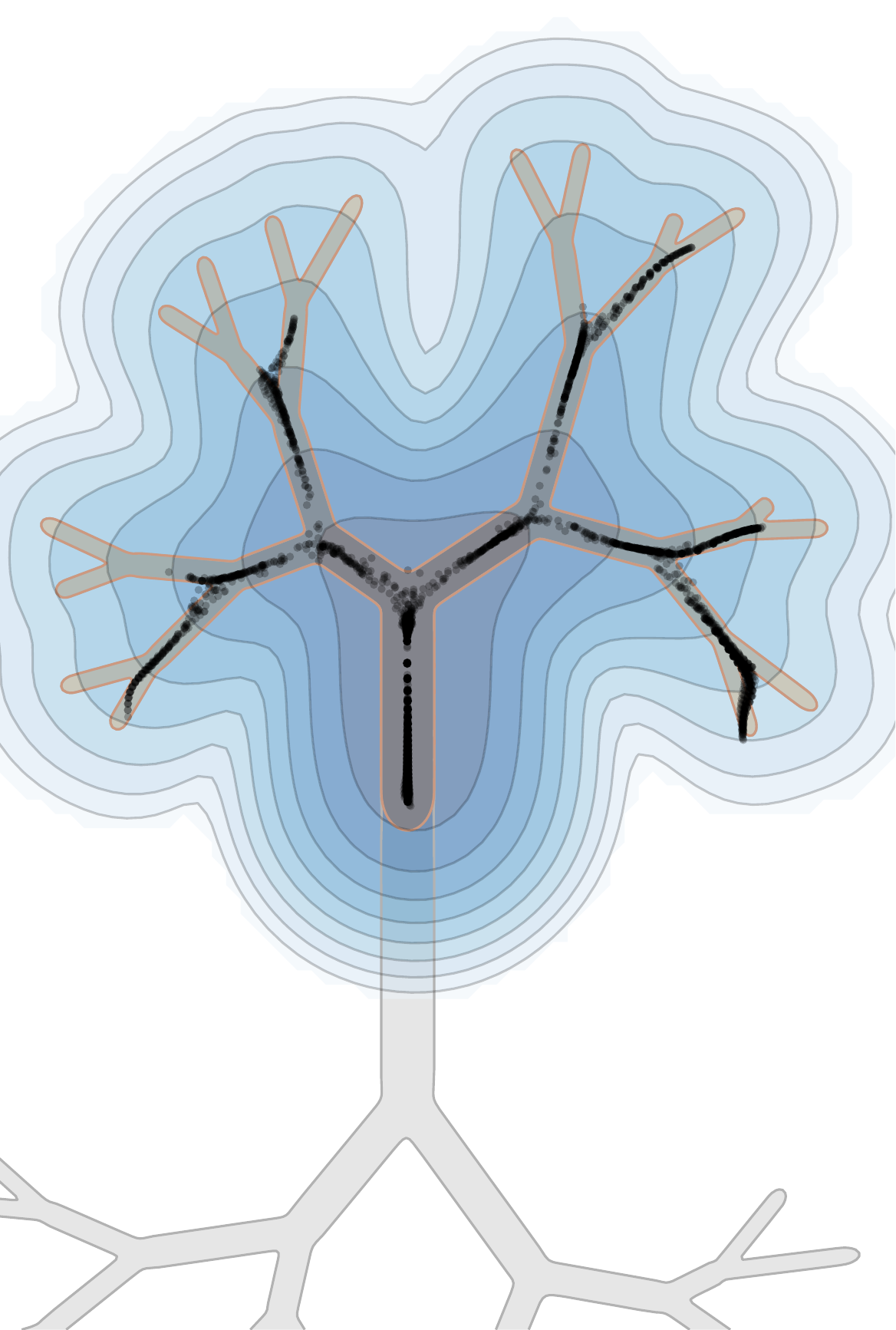}
        \\ 
        (a) Ground truth & (b) DDIM  & (c) SDS  & (d) \textbf{Our MSD}
        \\[1mm]
        \multicolumn{4}{c}{\datasample {\small \, data sample} $\quad$ \includegraphics[width=0.02\textwidth]{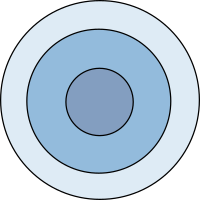} {\small \, loss landscape}}
    \end{tabular}
    \vspace{-2mm}
    \caption{
        Mode-seeking simulated in a fractal-like 2D distribution with two (\classone, \classtwo) classes, adapted from \cite{karras2024guiding}. We compare the behavior of diffusion sampling (DDIM) to optimization-based diffusion distillation, in a class-conditional setting. With class=\classone, \textbf{(a)} Ground truth distribution, \textbf{(b)} DDIM sampling , \textbf{(c)} SDS, and \textbf{(d)} our MSD. All methods are run without guidance.
    }
    \label{fig:fractal_2D_part1}
    \vspace{-5mm}
\end{figure*}


\section{Mean-Shift Distillation}
\label{sec:method}

In this section we derive the mean-shift vector for the diffusion output distribution, and show how it approximates the gradient thereof. We further show how an efficient estimate of this vector may be obtained with a minimal modification of diffusion sampling. We begin with a motivation of our development by illustrating the pitfalls of SDS.

\subsection{Motivation}
\label{sec:motivation}

Given a pre-trained diffusion model $\epsilon_\phi$, the SDS loss penalizes the KL-divergence of a unimodal Gaussian distribution centered around $x$ and the data distribution $p_\phi(z_t; c, t)$ captured by the frozen diffusion model conditioned on text embeddings $c$. With $x=g(\vartheta)$, an image rendered by $\vartheta$ via a differentiable renderer $g$, \cite{poole2022dreamfusion} derive the gradient of the loss $\mathcal{L}_\mathrm{SDS}$ with respect to $\vartheta$:
\begin{multline}
    \label{eq:vanilla_sds}
        \nabla_{\vartheta} \mathcal{L}_\mathrm{SDS} = \mathbb{E}_{
        t, \epsilon
        } \Big[
        \alpha(t)
        \left(
         \epsilon_{\phi}(\alpha(t)x+\epsilon;t) - \epsilon 
         \right)
         \Big] \frac{\partial x}{\partial \vartheta},\\[2mm]
         \mathrm{with}\;t\sim U(0,T),\epsilon \sim \mathcal{N}(0,\sigma(t)I).
\end{multline}
To illustrate the pitfalls of SDS, we simulate it in 2D using a small denoising diffusion network (\cref{fig:fractal_2D_part1}). This allows us to set $\vartheta = x \in \mathbb{R}^2$ (where $g$ becomes an identity map). 
We construct a fractal-like dataset as shown by \cite{karras2024guiding}, with analytic ground-truth probability density and score. 
This data distribution is a mixture of highly anisotropic Gaussians, where most of the probability mass resides in narrow regions, emulating the low intrinsic dimensionality of natural images \cite{Roweis2000, MikhailLaplace}. 
For a baseline, we compare it with DDIM \cite{song2021denoising}, a popular first-order sampling algorithm, with classifier-free guidance (CFG) \cite{ho2021classifierfree}. 
More details can be found in~\cref{sec:toydist_subsec}.

It is immediately apparent how even in this simple setting, the optima to which SDS converges do not model the output distribution well. Furthermore, the convergence itself is problematic due to very high variance of SDS, which we will address later.

\subsection{Mean-shift Gradient Approximation}
\label{sec:mean_shift}

\newcommand{\bandwidth}{\lambda}
\newcommand{\kernel}{{K_{\bandwidth}}}
\newcommand{\gaussian}{{G_{\bandwidth}}}
\renewcommand{\ps}{p^{\ast}_{\bandwidth}}
\newcommand{\psd}{\dot{p}_{\bandwidth}}

We start by convolving the data density $p$ with a radial Gaussian kernel $\gaussian(x)=c_\bandwidth e^{-x^2/\bandwidth^2}$ with bandwidth $\bandwidth$, normalized by a constant $c_\bandwidth$. This convolution yields a smoothed density $\ps(x)$:
\begin{equation}
\label{eq:gradient_smoothed_density}
    \ps(x) = p \ast \gaussian (x) = \int \gaussian(x-y) p(y) dy.
\end{equation}
We now take the gradient $\nabla_x \ps(x)$ of the smoothed density and substitute the Gaussian kernel's gradient:
\begin{align}
    \label{eq:distro_convolution}
    \nabla_x \ps(x)
        &= \int \nabla_x\gaussian(x-y) p(y) dy \\
        &= \int c_\bandwidth (x - y) \gaussian(x-y) p(y) dy.
\end{align}
We then take the stationary-point equation and reorganize it as a fixed-point iteration:
\begin{equation}
    \label{eq:fixed_point_iteration}
    \!\!\!\nabla_x \ps\!(x) = 0
    \;\;\Longrightarrow\;\;
    x' \!= \frac{\int y \gaussian(x-y) p(y) dy}{\int \gaussian(x-y) p(y) dy},\!
\end{equation}
where the constant $c_\bandwidth$ cancels out. We will discuss the practical estimation of the integrals in \cref{sec:product_sampling} below.

The iterative process in \cref{eq:fixed_point_iteration} is a continuous version of mean shift~\cite{ComaniciuM02Meanshift}.
We may turn this into gradient proxy with several desirable properties. Defining the mean-shift vector $\vec{m}(x) = x' - x$, it follows from \cref{eq:distro_convolution} that
\begin{equation}
    \vec{m}(x) \, \propto \, \nabla_x \ps (x) / \ps (x).
\end{equation}
Since the smoothed density $\ps(x)$ is always non-negative, $\vec{m}(x)$ is always aligned with its gradient $\nabla \ps (x)$. It is also aligned with the gradient of the true density $p$ as $\bandwidth \rightarrow 0$ (when such gradient exists). This means that a differential step along the vector $\vec{m}(x)$ will improve the likelihood of $x$, making this a good proxy for the kernel density estimation gradient. Furthermore, it implies that $\vec{m}(x)$ will be zero at the modes of $\ps (x)$, giving us a convergence criterion.

\newcommand{\inversion}{\textcolor{orange}{$f^{-1}$}}

\begin{figure*}[t]
    \centering
    \setlength{\tabcolsep}{0.7mm}
    \renewcommand{\arraystretch}{0}
    \begin{tabular}{c c c c}
        \multicolumn{4}{c}{\includegraphics[width=1.0\linewidth]{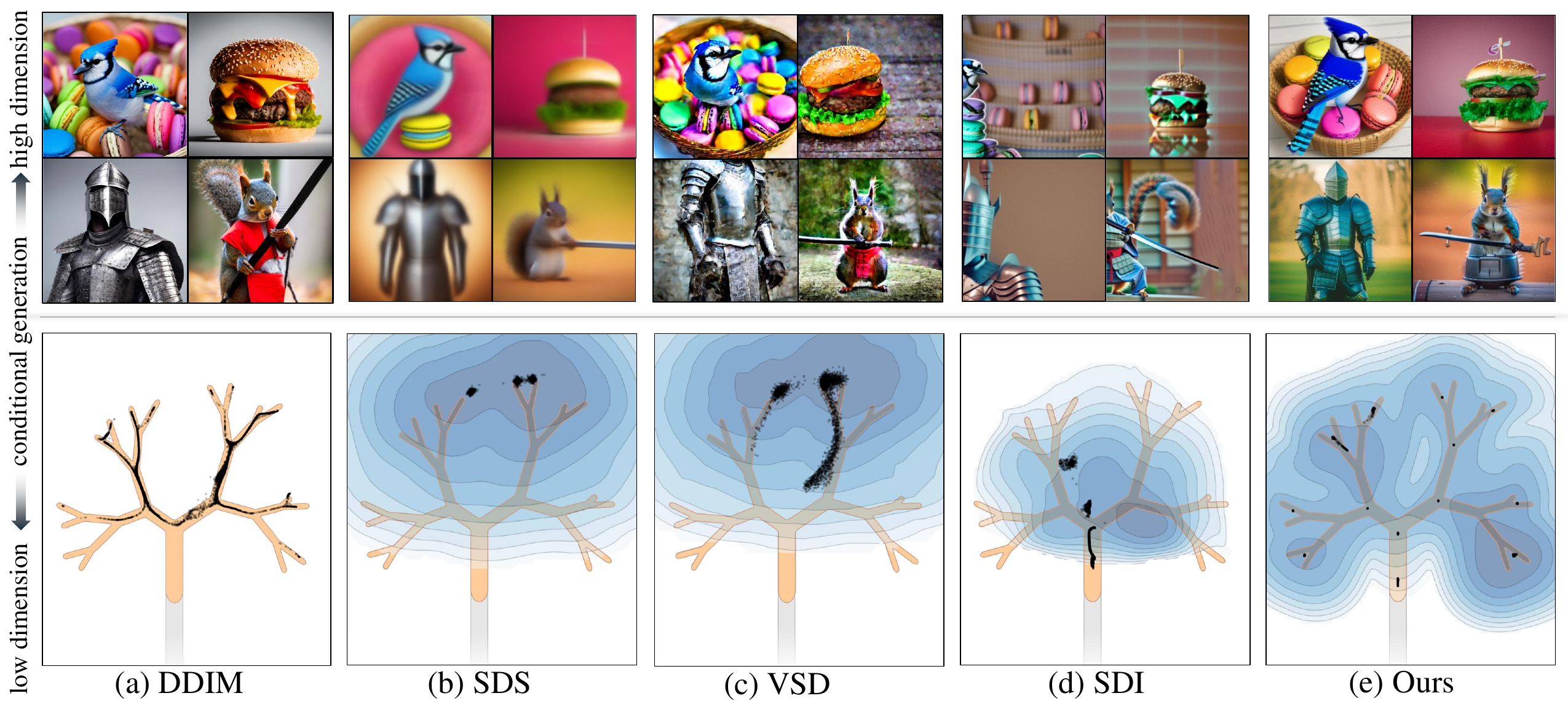}}
    \end{tabular}
    \caption{
        We juxtapose diffusion sampling vs diffusion distillation in low-dimensional ($\sR^2$) and high-dimensional ($\sR^{64 \times 64 \times 4}$) setting, using guidance via CFG \cite{ho2021classifierfree}. 
        \textbf{Top}: \textbf{(a)} text-conditioned generation of image via DDIM with 32 steps, \textbf{(b) - (e)} optimized coordinate-based neural implicit image for SDS, VSD, SDI, and our MSD respectively with StableDiffusion (CFG=7.5, §~\ref{sec:stablediffusion_exps}). \textbf{Bottom}: \textbf{(a)} class-conditioned generation of 2D points via DDIM with 32 steps, \textbf{(b) - (e)} optimized 2D points for SDS, VSD, SDI, and our MSD respectively (CFG=4, §~\ref{sec:toydist_subsec}). Text-prompts in clockwise order: \emph{``A DSLR photo of a ...} \emph{hamburger}, \emph{squirrel dressed as a samurai weighing a katana}, \emph{knight in silver armor}, and \emph{bluejay on basket of macarons"}.
    }
    \label{fig:fractal_2D_part2}
    \vspace{-4mm}
\end{figure*}


\subsection{Gradient Estimation via Product Sampling}
\label{sec:product_sampling}

The integrals in \cref{eq:fixed_point_iteration} can be both estimated using samples $y$ from the density $p$; such estimation yields the classical mean-shift expression
\begin{equation}
    \label{eq:mean_shift}
    x' = \frac{\sum_{y_i \sim p} G_\lambda (x - y_i) y_i}{\sum_{y_i \sim p} G_\lambda (x - y_i)}.
\end{equation}
In our case, we do not have such samples readily available. We could in theory use images from the training dataset as these samples, or else use the diffusion model to generate them -- either as a pre-process or on-the-fly during iteration. Unfortunately, that would be prohibitively costly as the datasets are typically quite large and accurate estimation would require a very large number of samples for practical (i.e.\ small) kernel bandwidths $\bandwidth$.

Our key insight is that the right-hand side of \cref{eq:fixed_point_iteration} can be viewed as an expectation with respect to a density $\psd$ that is the product of $p$ and the kernel $\gaussian$ centered at $x$:
\begin{equation}
    \label{eq:fixed_point_iteration_expectation}
    x' = \int y \, \psd(y|x) dy \,=\, \mathbb{E}_{y\sim \psd(y|x)}[y].
\end{equation}
To generate samples $y$ from this product density, we exploit the fact that diffusion models employ score-based sampling \cite{song2021scorebased, dhariwal2021diffusion}. Instead of using the score $\nabla\!\log p$ of the density $p$ in DDIM sampling, we use the score of our product density:
\begin{align}
        \nonumber
        \nabla_{\!z_t}\!\log(\psd\!(y|x))
            &= \nabla_{\!z_t}\!\log p(z_t) + \nabla_{\!z_t}\!\log \gaussian(x - z_t) \\
            &= \nabla_{\!z_t}\!\log p(z_t) - \frac{x - z_t}{\bandwidth^2},
    \label{eq:ms_guidance}
\end{align}
where $z_t = \alpha(t)x+\epsilon$. This is the sum of the density score (provided by the diffusion model) and the score of our Gaussian kernel. Having the ability to generate samples $y_i$ from the product density, we can estimate the mean-shift iterate (\ref{eq:fixed_point_iteration_expectation}) as
\begin{equation}
    \label{eq:mc_step_estimate}
    x' \approx \frac{1}{N} \sum_{y_i \sim \psd(y|x)} y_i.
\end{equation}
In practice we use a single sample $y$, which simplifies our mean-shift vector to 
\begin{equation}
\label{eq:mean_shift_vector}
    \vec{m}(x) = y - x.
\end{equation}
We can thus step along $\vec{m}$ to seek the modes of the data density~$p$. Substituting a learned score model into \ref{eq:ms_guidance} gives us
\begin{equation}
\label{eq:kernel_guided_score}
   \hat\epsilon_t = \epsilon_\theta(z_t;t) - \frac{x-z_t}{\lambda^2}.
\end{equation}

\subsection{Practical Considerations}

As noted in SJC \cite{sjc}, distillation with unconditioned diffusion models is challenging in high-dimensional settings like images. While we show unconditioned diffusion distillation is practical in simple 2D toy datasets below, we operate in the conditional setting throughout.

\paragraph{Impact of guidance.}

Conditional score estimates from diffusion models, $\epsilon_\theta (z_t, c) \approx - \sigma_t \nabla_{z_t} \log p(z_t | c)$, are improved in practice with classifier-free guidance (CFG) \cite{ho2021classifierfree}, which sharpens the distribution around the modes:
\begin{equation}
    \label{eq:cfg_update}
    \tilde{\epsilon}_\theta (z_t, c) = (1 + w) \epsilon_\theta (z_t, c) - w \epsilon_\theta (z_t).
\end{equation}
We may directly substitute this for the denoiser term in \cref{eq:kernel_guided_score}. 
Despite its practical success, the denoising direction induced by CFG does not provide theoretical guarantees in producing samples from $p_{0,w}(z_t|c)$ \cite{bradley2024classifierfreeguidancepredictorcorrector}. 
Even in simple settings, as observed in \cref{fig:fractal_2D_part1}(b), CFG can lead to mode drops. While alternative guidance strategy exists \cite{karras2024guiding}, we stick with the dominant practice of using CFG (\cref{eq:cfg_update}).
We have found that this synergizes well with mode-seeking by mean-shift, and show the effects of this in evaluation below. See discussion in Supplemental §B.

\begin{figure*}[t]
    \centering
    \includegraphics[width=0.97\linewidth]{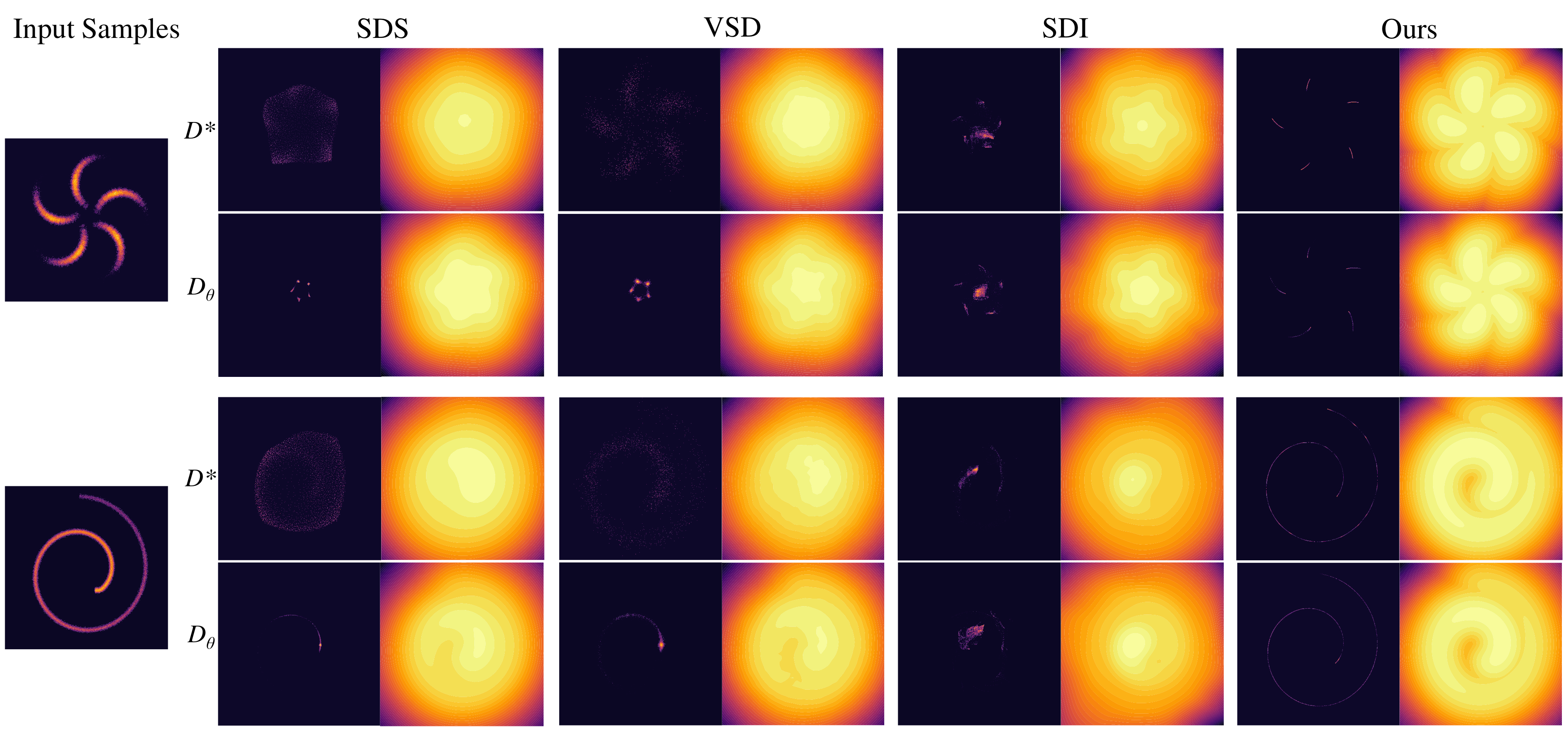}
    \vspace{-4mm}
    \caption{
    Unconditional distillation on two toy density datasets, \emph{Pinwheel} (top) and \emph{Spiral} (bottom), given an ideal denoiser (\textbf{$D^*$}) and a learned denoiser (\textbf{$D_\theta$}). For each method and denoiser, we show the optimized samples (left) and the loss landscape (right). \emph{Zoom in for clarity}.
    }
    \label{fig:toydensity}
    \vspace{-6mm}
\end{figure*}


\paragraph{Integrating kernel score.}
\label{sec:integrating}

Because the magnitude of the kernel term in \cref{eq:kernel_guided_score} can be quite high when \mbox{$|y-z_t|$} is high relative to $\lambda$, directly implementing this can result in instability while denoising particularly with explicit integrators. Higher-order integrators are generally capable of dealing with this instability, but require many more score function evaluations. 

To address this, we note that in isolation the kernel term has the form of a negative exponential centered on $y$, or explicitly:
\begin{equation}
    z_{t+\Delta t}=y + (z_t - y) e^{\frac{\Delta t}{\lambda^2}},
\end{equation}
where $\Delta t$ is negative. We take advantage of this to formulate a stable approximation that avoids the stability issues with a minimal change to the integration process. Instead of feeding the full composite score function to the integrator, in each time step we first integrate only the score function with the existing integrator to get ${z'}_{t+\Delta t}$. Immediately after, we separately account for the kernel term by computing the final output as
\begin{equation}
    \label{eq:integratingkernel}
    z_{t+\Delta t}=y+({z'}_{t+\Delta t} - y) e^{\frac{\Delta t}{\lambda^2}}.
\end{equation}
A full derivation is provided in the supplemental. We note such numerically instability has been observed when using high CFG values. A remedy is to apply guidance in a limited interval \cite{kynkaanniemi2024applying}. We leverage similar ad-hoc tricks by applying the kernel term in limited interval through the sampling chain.

\vspace{-2mm}
\section{Practical Implementation and Evaluation}

In this section, we construct synthetic examples on which we demonstrate that our proposed method behaves as theory predicts, alleviating the issues SDS exhibits even in these simple scenarios. We further explain the issues encountered when translating this theory into practice, and describe adaptations we designed to make our method work with real-world diffusion models, retaining desirable properties. We make comparisons with two strong baselines that improve the convergence and performance of SDS; SDI \cite{lukoianov2024score}, who propose a better noise term to reduce late-stage stochasticity, yet, retain the same gradient computation of SDS, and VSD \cite{wang2023prolificdreamer}, who propose to learn the variational score as opposed to assuming it to be a known analytic score like in SDS.

\subsection{Idealized Setting}
\label{sec:ideal}

In order to manage large data dimensionality as well as massive training datasets, diffusion in practice employs a trained neural network to represent the denoiser $D$. However, \cite{Karras2022edm, symmetryJihyeon2024} have identified an analytical solution to minimizing the denoiser error, the \emph{ideal denoiser}
$D^*(x;t)$:
\begin{equation}
\label{eq:idealdenoiser}
    D^*(x;t)=
    \frac{\sum_i u_i \mathcal{N}(x;u_i,\sigma(t))}
    {\sum_i \mathcal{N}(x;u_i,\sigma(t))},
\end{equation}
where $u_0 \dots u_n$ are samples in our training set. 
Attentive readers will notice that this is in fact the discrete mean shift formula \cite{ComaniciuM02Meanshift}, with training samples taking the place of data samples and noise magnitude $\sigma(t)$ taking place of the kernel bandwidth $\lambda$. This expression is feasible to compute in practice for small datasets, and by setting 
\begin{equation*}
    \!\!\!\!\epsilon^*_\phi (z_t;t) = - \frac{D^*(z_t;t) - z_t}{\sigma_t},
\end{equation*}
we may substitute it into the SDS formula (\ref{eq:vanilla_sds}) to get an explicit solution for the SDS gradient
\begin{equation*}
    \!\!\!\nabla_{x} \mathcal{L}_\mathrm{SDS} = \mathbb{E}_{t, z_t \sim \mathcal{N}(\alpha_tx,\sigma_t^2\mathbf{I})} \Bigg[ w(t)  \frac{z_t - D^*(z_t;\sigma_t)}{\sigma_t} \frac{\partial x}{\partial \theta} \Bigg].    
\end{equation*}
We can brute force numerically evaluate this integral. We can compare both methods on synthetic datasets, eliminating any error introduced by training and evaluating a neural model to show that the theoretical properties hold.


\newcommand{\RowHline}[1][4pt]{%
  \\\addlinespace[#1]\Xhline{0.1\arrayrulewidth}\addlinespace[#1]%
}


\begin{table}[t]
    \centering
    \small
    \caption{
    Metrics for class-conditional distillation on 2D Fractal dataset. For each metric, \emph{top} to \emph{bottom}: ideal denoiser (\textbf{$D^*$}), learned denoiser without guidance (\textbf{$D_\theta$} no CFG),  learned denoiser with CFG (\textbf{$D_\theta$} w/CFG) \cite{ho2021classifierfree}, and learned denoiser with Autoguidance (\textbf{$D_\theta$} w/AG)\cite{karras2024guiding}. \hlgold{$1^{\text{st}}$} and \hlsilver{$2^{\text{nd}}$} best among distillation-based methods for each column, highlighted. 
    }
    \vspace{-2mm}
    \label{tab:nll}
    \begin{tabular}
        {
        @{} l c@{\hspace{0mm}}
        *{1}{c}@{\hspace{0mm}}   
        *{1}{c}@{\hspace{0mm}}   
        *{1}{c}@{\hspace{0mm}}   
        *{1}{c}@{}   
        }
        \multicolumn{5}{@{\hspace{3mm}}c@{}}{Two-class Fractal dataset}\\
        \toprule
        \textbf{Method}
        & \multicolumn{1}{@{\hspace{3mm}}c@{}}{\textbf{NLL} $\downarrow$}
        & \multicolumn{1}{@{\hspace{3mm}}c@{}}{\textbf{MMD} $\downarrow$} 
        & \multicolumn{1}{@{\hspace{2mm}}c@{}}{\textbf{Precision} $\uparrow$}
        & \multicolumn{1}{@{\hspace{2mm}}c@{}}{\textbf{Recall} $\uparrow$} \\
        \midrule
        DDIM-\textbf{$D^*$} & -1.85 & 0.860  & 0.97 & 0.93 \\
        SDS-\textbf{$D^*$} & 36.15 & 328.0  & 0.08 & \hlsilver{0.03} \\
        VSD-\textbf{$D^*$} & \hlsilver{9.97} & 230.9  & 0.05 & 0.02 \\
        SDI-\textbf{$D^*$} & 24.28 & \hlgold{29.93}  & \hlsilver{0.27} & 0.01 \\
        \methodname-\textbf{$D^*$} & \hlgold{-1.32} &  \hlsilver{30.46} & \hlgold{0.92} &  \hlgold{0.33} 
        \RowHline 
        DDIM-\textbf{$D_\theta$} no CFG & -1.51 &  0.007 & 0.95 & 0.96 \\
        SDS-\textbf{$D_\theta$} no CFG & 9.12 &  \hlsilver{87.04} & 0.01 & 0.00 \\
        VSD-\textbf{$D_\theta$} no CFG & 9.88  & 94.68  & 0.10 & 0.05 \\
        SDI-\textbf{$D_\theta$} no CFG & \hlgold{-2.87} & 459.9 & \hlgold{0.97} & \hlsilver{0.12} \\
        \methodname-\textbf{$D_\theta$} no CFG & \hlsilver{-2.02}  & \hlgold{12.79}  & \hlgold{0.97} & \hlgold{0.42}
        \RowHline 
        DDIM-\textbf{$D_\theta$} w/CFG  & -1.59 & 257.43  & 0.97 & 0.44 \\
        SDS-\textbf{$D_\theta$} w/CFG  & \hlsilver{15.96} & 3875.11  & 0.17 & 0.03 \\
        VSD-\textbf{$D_\theta$} w/CFG  & 18.97 &  3845.41 & 0.21 & 0.05 \\
        SDI-\textbf{$D_\theta$} w/CFG  & 27.37  &  \hlgold{69.23} & \hlsilver{0.30} & \hlgold{0.48} \\
        \methodname-\textbf{$D_\theta$} w/CFG  & \hlgold{-1.15} & \hlsilver{133.41}  & \hlgold{0.94} & \hlsilver{0.40}
        \RowHline 
        DDIM-\textbf{$D_\theta$} w/AG & -1.67  & 0.25  & 0.96 & 0.79 \\
        SDS-\textbf{$D_\theta$} w/AG & 11.33  & \hlsilver{71.05}  & 0.04 & 0.03 \\
        VSD-\textbf{$D_\theta$} w/AG &  11.52 & \hlgold{70.25}  & 0.03 & 0.03 \\
        SDI-\textbf{$D_\theta$} w/AG & \hlsilver{0.65}  &  15089 & \hlsilver{0.51} & \hlgold{0.51} \\
        \methodname-\textbf{$D_\theta$} w/AG & \hlgold{-1.99}   &  122.94 & \hlgold{0.97} & \hlsilver{0.43} \\
        \bottomrule
    \end{tabular}
    \label{tab:2d_density_cond}
    \vspace{-3mm}
\end{table}


\begin{table}[t]
    \small
    \centering
    \caption{
       Metrics for unconditional distillation on 2D toy datasets. For each metric, \emph{top} to \emph{bottom}: ideal denoiser (\textbf{$D^*$}) and learned denoiser (\textbf{$D_\theta$}). MMD scaled by $10^{-4}$.
    }
    \vspace{-2mm}
    \label{tab:2d_density_uncond}
    \begin{tabular}{
        @{} c@{\hspace{0mm}}
        *{1}{c}@{\hspace{0mm}}   
        *{1}{c}@{\hspace{0mm}}   
        *{1}{c}@{\hspace{0mm}}   
        *{1}{c} @{} 
        }
        \multicolumn{5}{@{\hspace{3mm}}c@{}}{Unlabeled Spiral dataset}\\
        \toprule
        \textbf{Method} 
        & \multicolumn{1}{@{\hspace{3mm}}c@{}}{\textbf{NLL} $\downarrow$}
        & \multicolumn{1}{@{\hspace{3mm}}c@{}}{\textbf{MMD} $\downarrow$}
        & \multicolumn{1}{@{\hspace{2mm}}c@{}}{\textbf{Precision} $\uparrow$}
        & \multicolumn{1}{@{\hspace{2mm}}c@{}}{\textbf{Recall} $\uparrow$}\\
        \midrule
        DDIM-\textbf{$D^*$} & -1.39 & 0.410 & 0.97 & 0.93 \\ 
        SDS-\textbf{$D^*$} & 30.37 & \hlsilver{13.85} & 0.02 & 0.03 \\ 
        VSD-\textbf{$D^*$} & \hlsilver{10.15} & 23.46 & 0.04 & 0.09 \\
        SDI-\textbf{$D^*$} & 35.64  & 39.51 & \hlsilver{0.10} & \hlgold{0.90} \\ 
        \methodname-\textbf{$D^*$} & \hlgold{-1.28} & \hlgold{4.490} & \hlgold{0.99} & \hlsilver{0.18}
        \RowHline 
        DDIM-\textbf{$D_\theta$} & -1.32 & 1.160 & 0.96 & 0.96 \\ 
        SDS-\textbf{$D_\theta$} & \hlsilver{8.13} & 274.3 & 0.04 & 0.11 \\ 
        VSD-\textbf{$D_\theta$} & 8.90 & \hlsilver{271.8} & 0.07 & 0.14 \\
        SDI-\textbf{$D_\theta$} & 19.16 & 2008 & \hlsilver{0.12} & \hlgold{0.42} \\ 
        \methodname-\textbf{$D_\theta$} & \hlgold{-1.51} & \hlgold{18.41} & \hlgold{0.98} & \hlsilver{0.18} \\        
        \toprule
        \vspace{-2mm}
        \\
        \multicolumn{5}{@{\hspace{3mm}}c@{}}{Unlabeled Pinwheel dataset}\\
        \toprule
        \textbf{Method} 
        & \multicolumn{1}{@{\hspace{3mm}}c@{}}{\textbf{NLL} $\downarrow$}
        & \multicolumn{1}{@{\hspace{3mm}}c@{}}{\textbf{MMD} $\downarrow$}
        & \multicolumn{1}{@{\hspace{2mm}}c@{}}{\textbf{Precision} $\uparrow$}
        & \multicolumn{1}{@{\hspace{2mm}}c@{}}{\textbf{Recall} $\uparrow$}\\
        \midrule
        DDIM-\textbf{$D^*$} & -1.19 & 1.05 & 0.97 & 0.94 \\ 
        SDS-\textbf{$D^*$} & \hlsilver{2.29} & \hlgold{5.18} & \hlsilver{0.85} & \hlsilver{0.03} \\ 
        VSD-\textbf{$D^*$} & 3.34 & 6.78 & 0.65 & \hlgold{0.04} \\
        SDI-\textbf{$D^*$} & 28.31 & 6.13 & 0.17 & 0.001 \\ 
        \methodname-\textbf{$D^*$} & \hlgold{-1.94} & \hlsilver{5.83} & \hlgold{0.99} & 0.01
        \RowHline 
        DDIM-\textbf{$D_\theta$} & -1.1 & 0.270 & 0.97 & 0.97 \\ 
        SDS-\textbf{$D_\theta$} & \hlsilver{2.00} & 36.37 & 0.90 & 0.01 \\ 
        VSD-\textbf{$D_\theta$} & 2.28 & \hlsilver{33.36} & \hlsilver{0.97} & 0.02 \\
        SDI-\textbf{$D_\theta$} & 17.33 & 98.09 & 0.51 & \hlgold{0.15} \\ 
        \methodname-\textbf{$D_\theta$} & \hlgold{-2.19} & \hlgold{7.250} & \hlgold{0.99} & \hlsilver{0.13} \\
        \toprule
    \end{tabular}%
    \vspace{-4mm}
\end{table}

\subsection{Toy Distributions in \texorpdfstring{$\mathbb{R}^2$}{R\textsuperscript{2}}}
\label{sec:toydist_subsec}

In addition to the fractal dataset (\cref{fig:fractal_2D_part1}), we extend our analysis to other 2D datasets, with ${u_i}^m_{i=1} \subset \gM \in \sR^2$ sampled from various challenging toy 2D densities  \cite{scikit-image, rozen2021moserflowdivergencebasedgenerative}. For each, we sample $10^4$ points from the data distribution and initialize our target points densely across a grid $[-1.5, 1.5]^2$. With $10^3$ Monte Carlo samples, we benchmark SDS, VSD, SDI, and our \methodname using both an ideal denoiser (\ref{eq:idealdenoiser}) and a learned denoiser (\ref{eq:cfg_update}). For the fractal dataset the denoiser is class‑conditioned; its score is either left unchanged (without guidance) or guided via CFG or Autoguidance. For the other datasets, the denoiser is unconditioned.

We visualize the generated samples produced by all methods after the optimization in \cref{fig:fractal_2D_part2,fig:toydensity}. We also visualize the reconstructed loss functions. This makes the behavior of all methods particularly obvious; the peaks of this reconstructed function are out of distribution for all methods except our \methodname. Numerical evaluations in \cref{tab:2d_density_cond,tab:2d_density_uncond} show our method outperform baselines. We suspect that this bias persists in SDS in higher dimensional settings and is what causes SDS optimized results to be blurry and exhibit other artifacts (top row of \cref{fig:fractal_2D_part2}). More experimental details can be found in \cref{sec:evals} and the supplemental.

Beyond bias, we also track the gradient‑estimate variance, a key driver of convergence. Optimizing with stochastic gradients resembles a random walk: large variance slows progress, can push steps opposite the true gradient, and, once near an optimum, keeps the iterate oscillating instead of settling. To quantify the variance of an estimate $\hat{g}(x)$ of the gradient $g(x)$, we employ a slight variation of the Monte Carlo estimator efficiency formula 
$\varepsilon (\hat{g}(x))=|g(x)|^2 / \left(\mathrm{MSE}(\hat{g}(x)) \: \mathrm{cost}(\hat{g}(x)) \right)$. We measure cost as number of invocations of the score model, since that is the typical bottleneck in diffusion. 


Normalization by the squared norm of $g$ is included to account for the fact that, due to bias and scaling, different estimators may converge to gradients of different magnitude, and the normalized MSE then roughly describes the probability of the estimated gradient pointing the ``right'' way. MSE and cost are accumulated over many independent estimations, and average over many values of $x$.%

\noindent{}The result of these efficiency comparisons are in \cref{tab:efficiency} (in log-scale). Although getting a single estimate with our method requires more score model invocations, the efficiency of our method is significantly higher than SDS and VSD, and comparable to SDI. 

\begin{table}[t]
    \small
    \centering
    \vspace{0mm}
    \caption{
         Efficiency ($\uparrow$) on 2D toy density datasets. \emph{Left}: ideal denoiser / \emph{Right}: learned denoiser.
    }
    \vspace{-1mm}
    \label{tab:efficiency}
    \setlength{\tabcolsep}{1.5mm}
    \begin{tabular}{lccc} 
         \toprule
          & \textbf{Fractal} & \textbf{Spiral} & \textbf{Pinwheel}  \\
         \midrule
         SDS &  -7.37 / -6.89 & -8.48 / -7.57 & -7.82 / -6.99 \\
        \addlinespace[2pt]
         VSD &  -5.92 / -3.83 & -6.85 / -4.45 &  -6.36 / -3.93 \\
        \addlinespace[2pt]
         SDI & 14.18/ 14.21 &   13.94 / 14.17 &   14.19 / 14.18 \\
        \addlinespace[2pt]
        \methodname & 13.44 / 7.65 & 13.38 / 6.32 & 13.76 / 7.08 \\
        \bottomrule
    \end{tabular}
    \vspace{-3mm}
\end{table}

\subsection{Practical Setting}
\label{sec:practical_setting}

For large-scale image datasets, idealized denoiser is no longer tractable and we contend with a learned denoising function, and the associated machinery. This introduces numerical issues. Namely, the magnitude of the kernel term may grow to where the standard first or second order integrators can no longer manage it (\cref{sec:integrating}); but conversely, so does the magnitude of the learned score when $z_t$ is far out of distribution, because the ideal denoiser (\cref{sec:ideal}) uses the same equation as mean shift. Start of the optimization, it is likely in a high-dimensional space that $x$ will be out of distribution and we have to choose between the integration failing because the denoiser term has a high magnitude, or because the kernel term has a high magnitude.

To alleviate this, we use two heuristic approximations: applying guidance in limited interval (\cref{sec:integrating})
and scaling our sample in \cref{eq:integratingkernel} by noise corresponding to time step $t$. 
In practice, we apply inversion to get the latter. These are designed to keep the iterate in a region with reasonable score magnitude and still sample a distribution that is an approximation of the product distribution.

\subsection{Pre-trained Stable Diffusion}
\label{sec:stablediffusion_exps}

We use the latent-space diffusion model, Stable Diffusion, as the diffusion prior for text-conditioned optimization of parameters of differentiable image generators. Specifically, we optimize parameters $\vartheta$ of generator $g$, a rendering function that maps $\vartheta$ to an image $\mathcal{I}$. The rendered image $\mathcal{I}$ is fed to the image encoder to get $x^k$, our latent at optimization step $k$, over which the gradient is computed. We define two settings where $\vartheta$ (1) represents an RGB image, and (2) represents a 3D volume. Specifically:
\begin{enumerate}
    \item \textbf{Text-to-2D.} We represent 2D images via a coordinated-based MLP $f$ with learnable parameters $\vartheta$ that takes as input a 2D point $p$ in the unit square $p=(x,y) \in [0,1]^2$ and outputs RGB $ \in [0,1]^3$; $f(p;\vartheta):\mathbb{R}^2 \rightarrow $ RGB. We use this non pixel-based representation of an image for two reasons, (1) to prevent our method and the baselines from taking the exact gradient step i.e. running diffusion sampling and setting $x^k$ to the denoised latent $z_0$, and (2) we can directly compare with images sampled via DDIM, an unconstrained image generation setting.
    \item \textbf{Text-to-3D.} We represent 3D volumes as NeRFs, following \cite{poole2022dreamfusion}. The NeRF is parameterized by two MLPs, one for foreground and one for background. The former has 64 hidden nodes and 2 layers, with input $(x,y,z)$ coordinates encoded via HashGrid \cite{mueller2022instant}.
\end{enumerate}

\vspace{-4mm}
\paragraph{Implementation details.}

We implement all our code in PyTorch, on a single NVIDIA A100 gpu. We use the Threestudio \cite{threestudio2023} framework for experiments involving pre-trained Stable Diffusion. We use AdamW optimizer with lr$=10^{-2}$. We set optimization steps to $400$ for text-to-2D and $10k$ for text-to-3D. We use a monotonically decreasing schedule for the bandwidth $\lambda$.

\begin{table}[t]
    \small
    \centering
    \captionof{table}{%
       Text-to-2D quantitative comparison. We evaluate fidelity with FID and CLIP-SIM. $^\dagger$FID measured with DDIM as ground truth.
    }
    \vspace{-4mm}
    \label{tab:image_fusion}
    \setlength{\tabcolsep}{2mm}
    \begin{tabular}{@{}lcc@{}}
      \toprule
      \textbf{Method} & \textbf{FID} $\downarrow$ & \textbf{CLIP-SIM (L/14) $\uparrow$} \\
      \midrule
      DDIM$^\dagger$ & –      & 44.1$\pm$ 2.8 \\
      SDS            & 198.90 & 27.7$\pm$ 1.9 \\
      VDS            & 130.22 & 30.8$\pm$ 1.4 \\
      SDI            & 166.16 & 31.0$\pm$ 0.7 \\
      \methodname    & 114.12 & 32.6$\pm$ 0.8 \\
      \bottomrule
    \end{tabular}
    \vspace{-2mm}
\end{table}

\newcommand{\im}[1]{\includegraphics[width=0.2\linewidth]{#1}}

\begin{figure*}[t]
    \centering
    \setlength{\tabcolsep}{0pt}
    \includegraphics[width=0.93\linewidth]{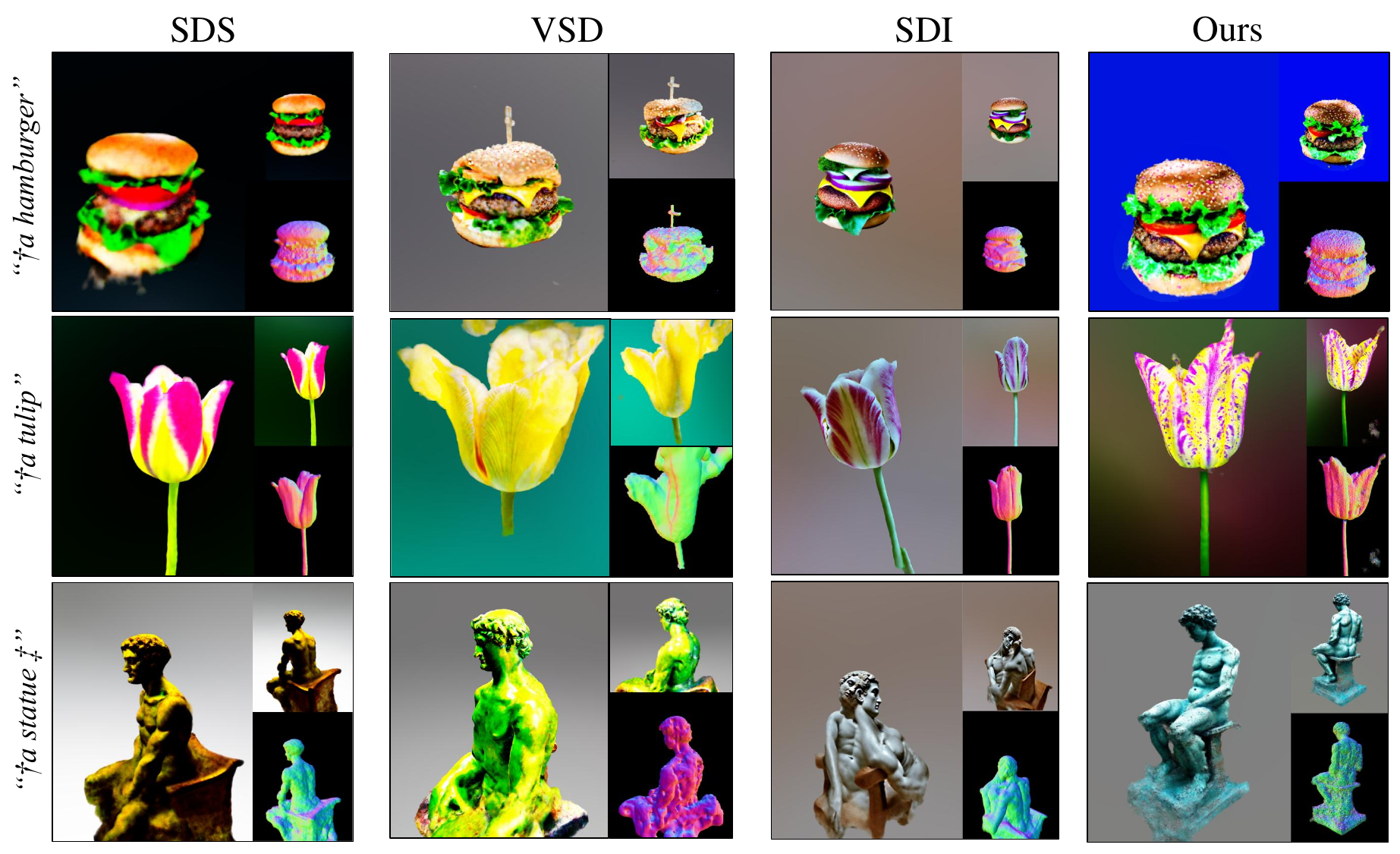}
    \vspace{-2mm}
    \caption{
        Comparison of 3D generation with other score distillation methods. Full prompt: $^\dagger$\emph{``A DSLR photo of a ..."}, $^\ddagger$\emph{``a Michelangelo statue of a man on a chair"}.}
    \label{fig:textto3d}
\vspace{-4mm}
\end{figure*}


\subsection{Evaluation}
\label{sec:evals}

\paragraph{Dataset.}

We use a subset of the prompts curated by \cite{poole2022dreamfusion,hertz2023delta}. We include all prompts in the supplemental.

\vspace{-3mm}
\paragraph{Metrics.}

For toy density dataset (\cref{sec:toydist_subsec}), we compute negative log-likelihood scores (NLL), generative precision and recall~\cite{Kynkaanniemi2019}, and maximum mean discrepancy (MMD).
For text-to-2D, we use images produced by DDIM to represent the ground truth distribution. To evaluate fidelity of the images, FID~\cite{heusel2017fid} is computed for each baseline (SDS, VSD, SDI) and ours against this ground truth image set, across 10,000 images. We also compute CLIP scores~\cite{clipscore} to measure prompt-generation alignment.

\vspace{-3mm}
\paragraph{Quantitative comparisons.}

\Cref{tab:image_fusion} reports results for FID and CLIP-based similarity, comparing our method with SDS, VSD, and SDI. We outperform all baselines in image fidelity and achieve faster convergence, as measured via \emph{fid} vs \emph{iterations} (see supplemental).

\vspace{-3mm}
\paragraph{Qualitative comparisons.}

\Cref{fig:fractal_2D_part2} (top row) and supplemental figure 3 compares our method with SDS, VSD, and SDI on text-to-2D generation, qualitatively. We show the importance of the two heuristics (\cref{sec:practical_setting}) to resolve numerical instabilities, absence of which can result in visual artifacts in supplemental figure 4. SDS, as discussed, produces low-fidelity results while SDI's inversion accumulates numerical errors during early stages of optimization.
In \cref{fig:textto3d}, we qualitatively compare results for text-to-3D optimization. We restrict to qualitative comparison for this task as quantitative metrics have high variance due to the absence of a ground truth dataset.

\begin{figure}[h]
  \vspace{-1mm}
  \centering
  \includegraphics[width=\linewidth]{
  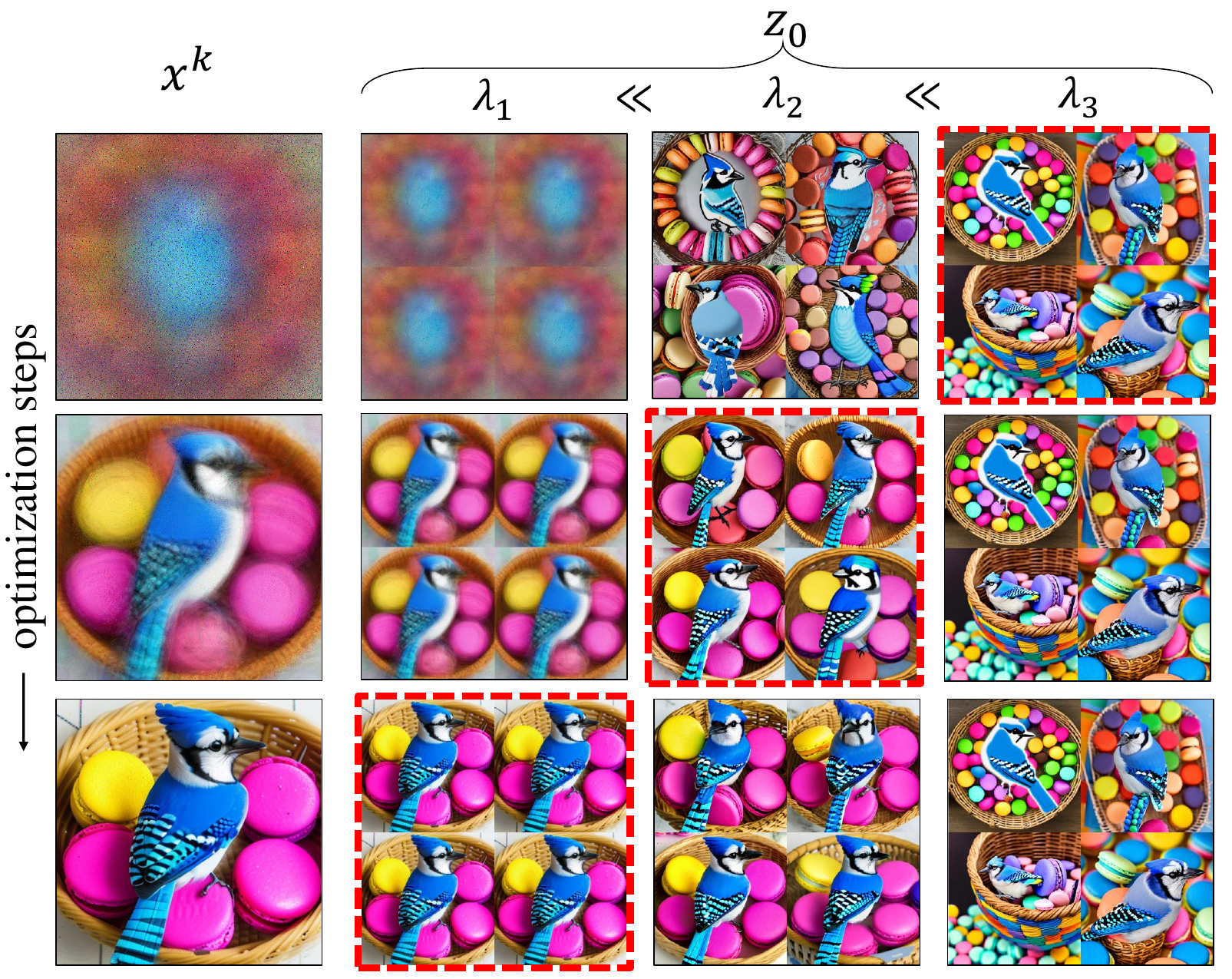
  }
  \captionof{figure}{
     Impact of bandwidth ($\lambda$) on the denoised latent ($z_0$). We set $\lambda_3 = 10^3$, $\lambda_2 = 10$, $\lambda_1 = 10^{-2}$. \textcolor{red}{Highlighted} images show the optimal bandwidth value corresponding to the $k^{th}$ optimization.
    }
  \label{fig:impact_of_lambda}
  \vspace{-5mm}                   
\end{figure}

\vspace{-4mm}
\paragraph{Impact of bandwidth.}

\Cref{fig:impact_of_lambda} shows the impact of the bandwidth ($\lambda$) term on the denoising process. First, we sample three parameters $\{\vartheta^k\}$ from our text-to-2D optimization pipeline at iterations $k=\{100, 200, 400\}$ and also sample three discrete $\lambda$ values $\{\lambda_1 \ll \lambda_2 \ll \lambda_3 \}$. Then, we run our forward pass once for each $\lambda_i$, independently. We visualize four decoded denoised latents $z_0$ (with different random seeds). The highlighted images show the optimal choices of $\lambda$ for each $x^k$ (the encoded latent for $\vartheta^k$). At high bandwidth value $\lambda_3$, the influence of the kernel term in the product sampling is negligible. This degenerates to vanilla denoising and we observe high variance in the output, irrespective of our current $x^k$. This is ideal at early stages of optimization. As bandwidth is annealed, we observe reduction in variance. Yet, the quality of the outputs can degrade if the kernel term dominates while $x^k$ is not \textit{``in-distribution''}. As $x^k$ approaches the mode of the distribution corresponding to the input text-prompt at final stages of optimization (when $k=400$), with a low bandwidth $\lambda_1$, our denoised latent $z_0 \approx x^k$. This provides us with a convergence criteria and we terminate when $\lambda$ is below the threshold $\lambda_1$.

\vspace{-1mm}
\section{Conclusion}

In this paper, we have reframed diffusion distillation in terms of explicitly ascending the gradient of the data distribution. We have derived mean-shift distillation as a proxy that provably aligns with this gradient, and in the limit its maxima are collocated with the modes of the data distribution. We have demonstrated that compared to SDS, this method achieves better mode alignment as well as lower gradient variance, which in practice translates to more realistic optimization results as well as improved convergence rate. Since this method simply provides optimization gradient much like SDS does, it may be used as a one-to-one replacement without retraining of the underlying model, or indeed substantial code modification. While the basic algorithm works as the theory predicts in synthetic scenarios, with real-world models we have to contend with integrator error due to large score magnitudes. We have designed heuristics to alleviate this and achieve improvements on SDS in practice, but we hope future work will be able to improve the integration and/or sampling procedure, obviating the need for heuristics, in-addition to using adaptive bandwidth annealing strategies.



\vspace{-2mm}
\section*{Acknowledgments}

The authors would like to thank Pradyumn Goyal, Dmitry Petrov, Ashish Singh, and Artem Lukoianov for their helpful discussions and insights. This project has received funding from the European Research Council (ERC) under the Horizon research and innovation programme (Grant agreement No. 101124742).

{
    \small
    \bibliographystyle{ieeenat_fullname}
    \bibliography{main}
}

\newpage
\onecolumn

\appendix
\section*{\centering {\Large Appendix}}

\section*{A: Discussions}
\label{sec:discussions}

\paragraph{Why mode-seeking?} The desirability of mode seeking varies between applications. When trying to directly sample images from the trained model, we wish to sample from the full variety of the distribution instead of getting only the mode--- we want sampling to interpolate between mode-seeking and mode-covering. Methods like DDIM aim for this. On the other hand, when we are optimizing an image (or using the image as a proxy to optimize, e.g. NeRF parameters), any gradient-based optimization will converge to a set of sparse points - local extrema - where the gradients are zero (if it converges at all). This is the intended use-case for SDS, VSD, SDI, and our method, and in this case, it is not possible in general to have the optimization process converge to a distribution of points. Given that, the best we can guarantee is that the points the process converges to are aligned with the distribution. Mode-seeking is our proposed way of achieving that.

\vspace{-2mm}
\paragraph{Compatibility with other guidance schemes.} 
In low-dimensional settings (eg, our toy experiments), our method can recover the modes and reconstruct the data distribution well without any guidance (See Figure 1 and 3). This is aided by the fact that the conditional score estimates parameterized as $\epsilon_\theta(z_t, c)$ (predicted noise from the pre-trained network) is good by itself, without guidance i.e. $\tilde{\epsilon}_\theta (z_t, c)$. Empirically, we observe that without guidance, ancestral sampling techniques like DDIM produce samples that lie on the data manifold, albeit with few outliers.

This is not the case in the high-dimensional setting with experiments on Stable Diffusion. Here $\epsilon_\theta(z_t, c)$ samples are noticeably bad and are predominantly outliers. Currently, the best fix is to augment these noise estimates with guidance to produce $\tilde{\epsilon}_\theta (z_t, c)$, the strategy prevalent in sampling algorithms. We inherit these practices when performing distillation.

Guidance mechanisms alternative to CFG \cite{ho2021classifierfree} have been proposed, like Autoguidance \cite{karras2024guiding}. As these methods pair well with DDIM (and other ancestral sampling techniques), we believe the benefits will extend to distillation-based methods like ours. Ultimately from the perspective of distillation, different guidance simply changes the shape of the output distribution but does not fundamentally change the mechanics of diffusion. While we show results using Autoguidance in Table 1, we used CFG in all the remaining experiments as it is more widely used, has hyperparameters (guidance scale) that have been more rigorously tested by the community, and was used in all our baselines (SDS, VSD, and SDI).

\vspace{-1mm}
\section*{B: Implementation Details}

\paragraph{Toy distributions in $\mathbb{R}^2$.}

For the learned denoiser, we use the architecture and training setup used by \cite{karras2024guiding} and similarly represent the densities as mixtures of Gaussians. 
We use the Adam optimizer \cite{kingmaB14Adam} and run the optimization procedure for 150 steps with a learning rate of $0.08$.
For our \methodname, we set an initial bandwidth of $0.316 \sim \sqrt{0.1}$ which is linearly decayed over the course of the optimization. 
For the ideal denoiser, due to it's high cost requirements in time and memory, we instead opt to use a few steps of gradient descent with high learning rate.
In addition to optimizing samples, we evaluate both SDS, VSD, SDI, and our \methodname gradients across the domain and then numerically integrate them to reconstruct the loss functions they represent.

\begin{figure}[h]
    \centering
    \includegraphics[width=0.5\linewidth]{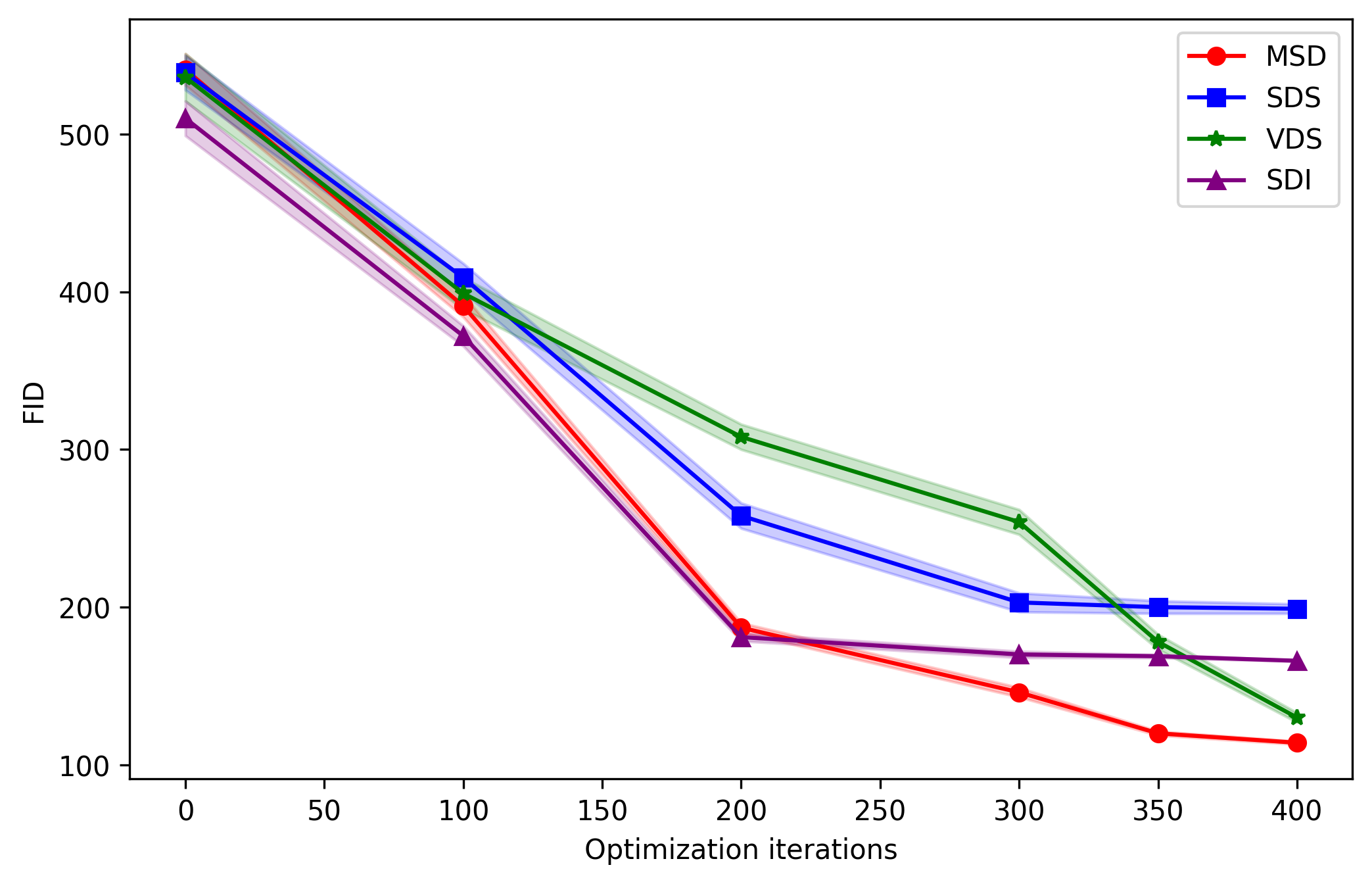}
\vspace{-2mm}
    \caption{FID vs optimization iterations for text-to-2D generation.}
    \label{fig:fid_vs_optimization}
\vspace{-3mm}
\end{figure}

\paragraph{Runtime and memory usage.} For text-to-2D generation, we report FID vs number of optimization iterations in \cref{fig:fid_vs_optimization}, extending the results in Table 4. Our method converges faster when compared to SDS, VSD, and SDI. The 3D NeRF training stage consumes 27GB GPU memory with 512 rendering resolution and batch size of 1.

\paragraph{Pseudocode.} We provide detailed pseudocode, comparing baseline SDS and our \methodname in ~\cref{alg:sds_vs_ours}.


\begin{figure*}[]
\SetKwInOut{Input}{Input}
\SetKwInOut{Output}{Output}
\Input{pre-trained diffusion model $\epsilon_\theta : \sR^{d_1 \times \dots \times d_k} \to \sR^{d_1 \times \dots \times d_k}$, target parameters $\psi \in \sR^d$, condition $c$, mapping function $g(\psi) : \sR^d \to \sR^{d_1 \times \dots \times d_k}$, time-dependent functions $w(t), \alpha(t)$, Monte Carlo sample size $N$.\\
}
\Output{$\psi^{*}$}
\begin{minipage}[t]{.43\textwidth}
    \centering
    \begin{algorithm}[H]
    \caption{Distillation via SDS}
    \label{alg:sds}
    \For{$k = 1,\hdots,steps$}{
        $x^k \gets g(\vartheta)$\\
        \For{$i = 1,\hdots,N$}{
            $t  \gets \text{U}(0, 1)$\\
            $z_t \gets \alpha(t)  x^k + \epsilon_t$\\
            $y_i \gets w(t)  [ \epsilon_\theta(z_t, t,c) - \epsilon_t ]$
        }
        $\nabla_{\vartheta} \mathcal{L}_{SDS} \gets \frac{1}{N} \sum (y_i - x^k)$\\
        
        \Comment{\textcolor{commentgray}{Backprop $\nabla_{\vartheta} \mathcal{L}_{SDS}$, update $\vartheta$}}
    }
    \end{algorithm}
\end{minipage}
\hspace{0.7cm}
\begin{minipage}[t]{.5\textwidth}
    \centering
    \begin{algorithm}[H]
    \caption{Distillation via \methodname (Ours)}
    \label{alg:msg}
    \Function{ODESolver($x, \lambda$) (eq 12)}{
        $z_T \gets \N(0,I)$\\
        \For{$t = T,\hdots,1$}
        {
            $z_{t-1} \gets \epsilon_\theta(z_t, t, c) - (x - z_t) / \lambda^2$\\
        }
        return $z_0$
    }
    \Function{ODESolver($x, \lambda$, stable) (eq 15)} {
        $\{z^*_t\}^T_{t=1} \gets inversion(x)$\\
        $z_T \gets  z^{*}_T  +  (\epsilon - z^{*}_T) \expkernel $\\
        \For{$t = T,\hdots,1$}
        {
            $z_{t-1} \gets z^{*}_t +  (\epsilon_\theta(z_t, t, c) - z^{*}_t) \expkernel $\\
        }
        return $z_0$\\
    }
    \Comment{\textcolor{commentgray}{initialize $\lambda$, set $\lambda_{min}$}}
    \For{$k = 1,\hdots,steps$}{
        $x^k \gets g(\vartheta)$ \\        
        \For{$i = 1,\hdots,N$}{
            $y_i \gets \text{ODESolver}(x^k, \lambda)$
        }
        $\nabla_{\vartheta} \mathcal{L}_{\methodname} \gets \frac{1}{N} \sum^N_i (y_i - x^k)$\\
        
        \Comment{\textcolor{commentgray}{Backprop $\nabla_{\vartheta} \mathcal{L}_{\methodname}$, update $\vartheta$}}
        \Comment{\textcolor{commentgray}{Anneal $\lambda$}}
        \If{$\lambda < \lambda_{min}$}{
            \Comment{\textcolor{commentgray}{terminate}}
        }
    }
    \end{algorithm}
\end{minipage}
\caption{
    Pseudocode of SDS and our procedure, \methodname. We additionally show the numerically stable solver, \emph{ODESolver($\dots$, stable)}, which is used for experiments with Stable Diffusion. Note, there is stochasticity in the \emph{ODESolver}.
}
\label{alg:sds_vs_ours}
\end{figure*}

\section*{C: Integrating Kernel Term}

We derive the numerically stable update for the kernel term in equation 12 of the main paper. Consider only the kernel term in equation 12 centered on $y$:

\begin{equation*}
s(z_t) = \frac{y - z_t}{\lambda^2}
\end{equation*}
Expressing this as a first-order linear ODE, we get

\begin{equation*}
\frac{d z(t)}{dt} + \frac{z(t)}{\lambda^2} = \frac{y}{\lambda^2}
\end{equation*}
Multiply both sides by $e^{\frac{ t}{\lambda^2}}$

\begin{equation}
\label{appendix:eq_1}
\frac{d z(t)}{dt} e^{\frac{ t}{\lambda^2}} + \frac{z(t)}{\lambda^2} e^{\frac{ t}{\lambda^2}} = \frac{y}{\lambda^2} e^{\frac{ t}{\lambda^2}}
\end{equation}
The left-hand side of eq~\ref{appendix:eq_1} is the derivative of $e^{\frac{ t}{\lambda^2}} z(t)$:

\begin{equation*}
\frac{d }{dt} \left[ e^{\frac{ t}{\lambda^2}} z(t) \right] = \frac{y}{\lambda^2} e^{\frac{ t}{\lambda^2}}
\end{equation*}
Now, we can integrate both sides from $0$ to $t$ for continuous variable $t$ and solve for $z(t)$:
\begin{align*}
\int^t_0  \frac{d }{d\tau} \left[ e^{\frac{ \tau}{\lambda^2}} z(\tau) \right]  &= y \int^t_0  \frac{e^{\frac{ \tau}{\lambda^2}} }{\lambda^2} d \tau\\
e^{\frac{ t}{\lambda^2}} z(t) - z(0) &= y e^{\frac{ t}{\lambda^2}} - y\\
z(t) &= y + (z(0) - y) e^{\frac{-t}{\lambda^2}} 
\end{align*}
Advancing one discrete reverse step $\Delta t < 0$ starting at $z_t$
\begin{equation}
z_{t+\Delta t}=y + (z_t - y) e^{\frac{\Delta t}{\lambda^2}}
\end{equation}

\noindent This gives us the closed-form update shown in equation 14 of the main paper that moves the current iterate $z_t$ exponentially toward $y$ with time constant $\lambda^2$. We do not need a numerical integrator, resolving issues when magnitude of the kernel term is high.

\section*{D: Ablations and More Results}

\begin{figure*}[t]
    \centering
    \includegraphics[width=0.98\linewidth]{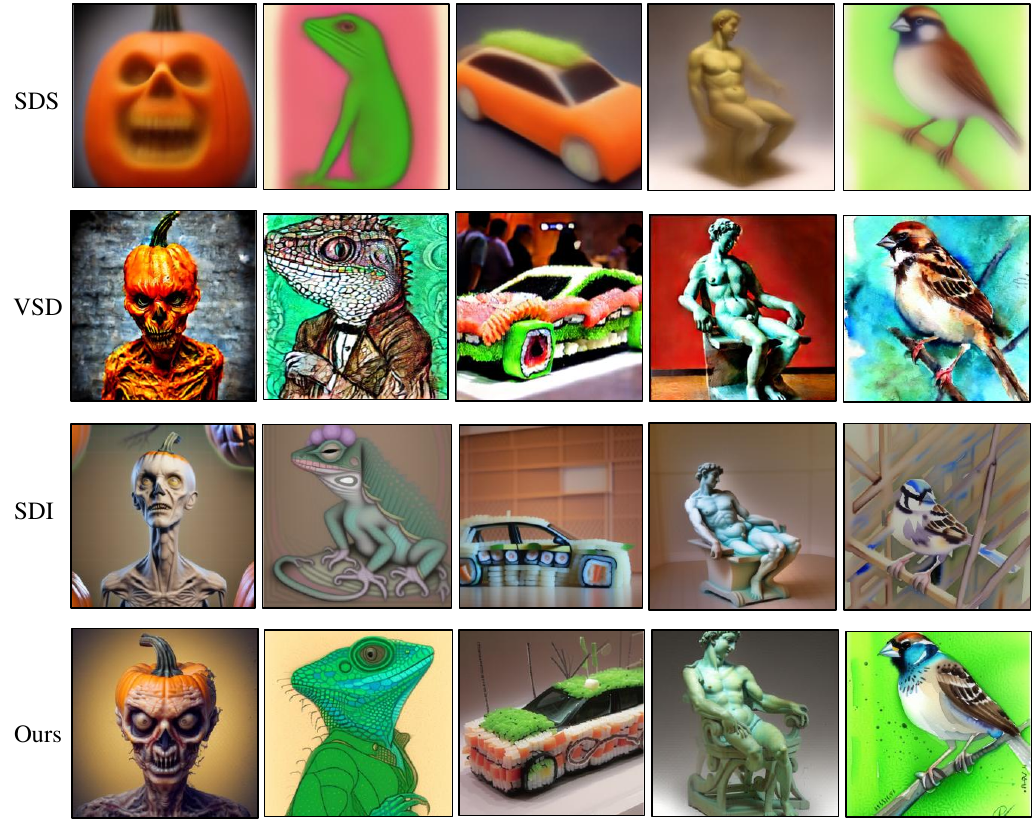}
    \vspace{-2mm}
    \caption{We also show additional results for our method (full), SDI, VSD, and SDS.}
    \label{fig:appendix_results_2}
    \vspace{-3mm}
\end{figure*}

\begin{figure*}[h]
    \centering
    \includegraphics[width=0.98\linewidth]{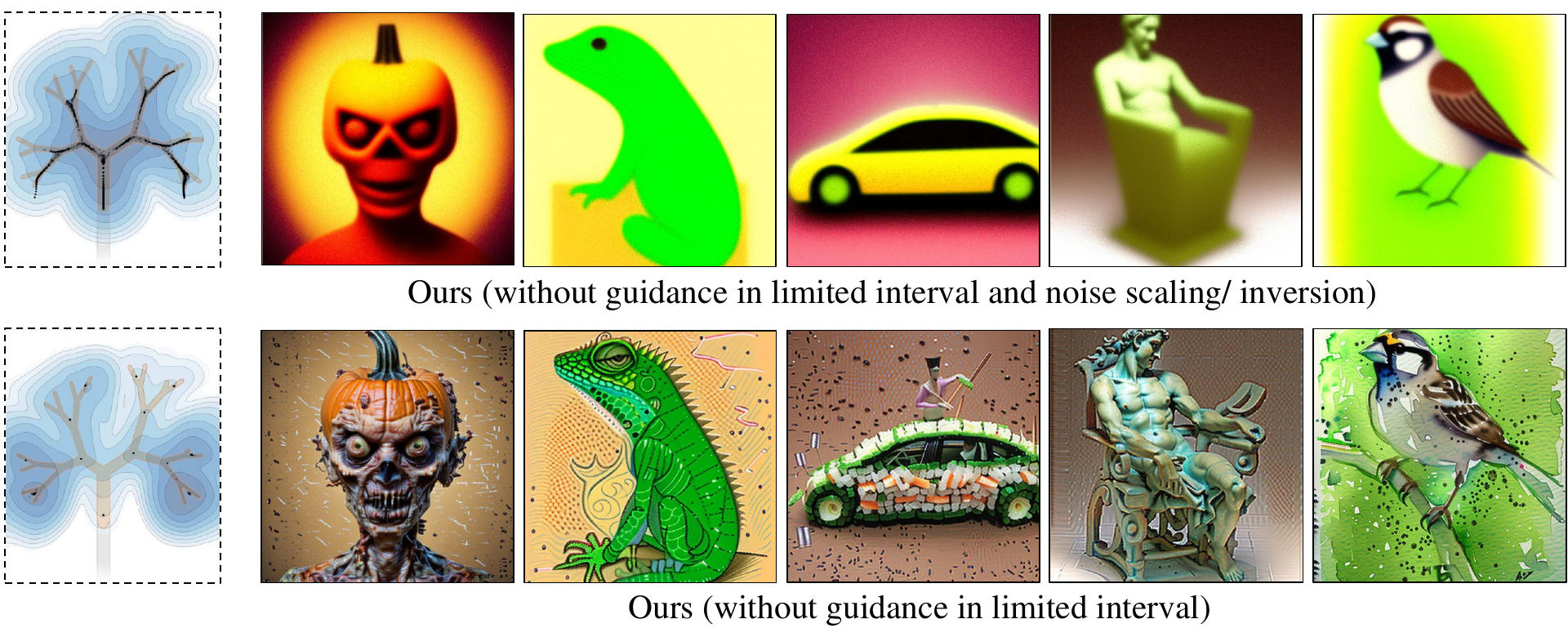}
    \vspace{-2mm}
    \caption{We extend Fig. 2 in the main paper with two ablations; applying guidance in the entire denoising trajectory (row 1) and noise scaled sample in the kernel term (row 2).}
    \label{fig:appendix_results_1}
    \vspace{-3mm}
\end{figure*}

We provide additional results in ~\cref{fig:appendix_results_2} and ~\cref{fig:appendix_results_1}.

\section*{E: List of Prompts}
\label{sec:prompts}

\begin{enumerate}
    \renewcommand{\labelenumi}{}
    \item \textit{``A DSLR photo of a hamburger''}
    \item \textit{``A blue jay standing on a large basket of rainbow macarons''}
    \item \textit{``A DSLR photo of a squirrel dressed as a samurai weighing a katana''}
    \item \textit{``A DSLR photo of a knight in silver armor''}
    \item \textit{``Line drawing of a Lizard dressed up like a victorian woman, lineal color''}
    \item \textit{``A photo of a car made out of sushi''}
    \item \textit{``A DSLR photo of a tulip''}
    \item \textit{``A DSLR photo of a Pumpkin head zombie, skinny, highly detailed, photorealistic''}
    \item \textit{``A watercolor painting of a sparrow, trending on artstation''}
    \item \textit{``Michelangelo style statue of man sitting on a chair''}
\end{enumerate}

\section*{F: Licenses}
Here we provide the URL, citations and licenses of the open-sourced assets we use in this work.

\begin{enumerate}
    \item \small{\url{https://github.com/threestudio-project/threestudio}} [Apache License 2.0].
    \item \small{\url{https://github.com/Stability-AI/stablediffusion}} [MIT License].
    \item \small{\url{https://github.com/NVlabs/edm2}} [CC BY-NC-SA 4.0].
\end{enumerate}

\end{document}